%% file: weakseg-cvpr.tex
\documentclass[10pt,twocolumn,letterpaper]{article}

\usepackage{cvpr}
\usepackage{times}
\usepackage{epsfig}
\usepackage{graphicx}
\usepackage{amsmath}
\usepackage{amssymb}
\usepackage{stfloats}

\usepackage{multirow}
\usepackage{booktabs} 

\usepackage{caption}

\usepackage{xcolor}

\usepackage[pagebackref=true,breaklinks=true,letterpaper=true,colorlinks,bookmarks=false]{hyperref}

 \cvprfinalcopy 

\def\cvprPaperID{4740} 

\ifcvprfinal\fi
\begin{document}

\title{Style Transfer by Relaxed Optimal Transport and Self-Similarity}

\author{\begin{tabular}{ccc}Nicholas Kolkin$^1$ & Jason Salavon$^2$ & Gregory %
                                                                      Shakhnarovich$^1$\end{tabular}\\%
$^1$Toyota Technological Institute at Chicago\hspace{5em}$^2$University of Chicago\\%
{\tt\small nick.kolkin@ttic.edu, salavon@uchicago.edu, greg@ttic.edu}}


\maketitle

\begin{figure*}[!b]
	\centering
	\includegraphics[width=0.95\textwidth]{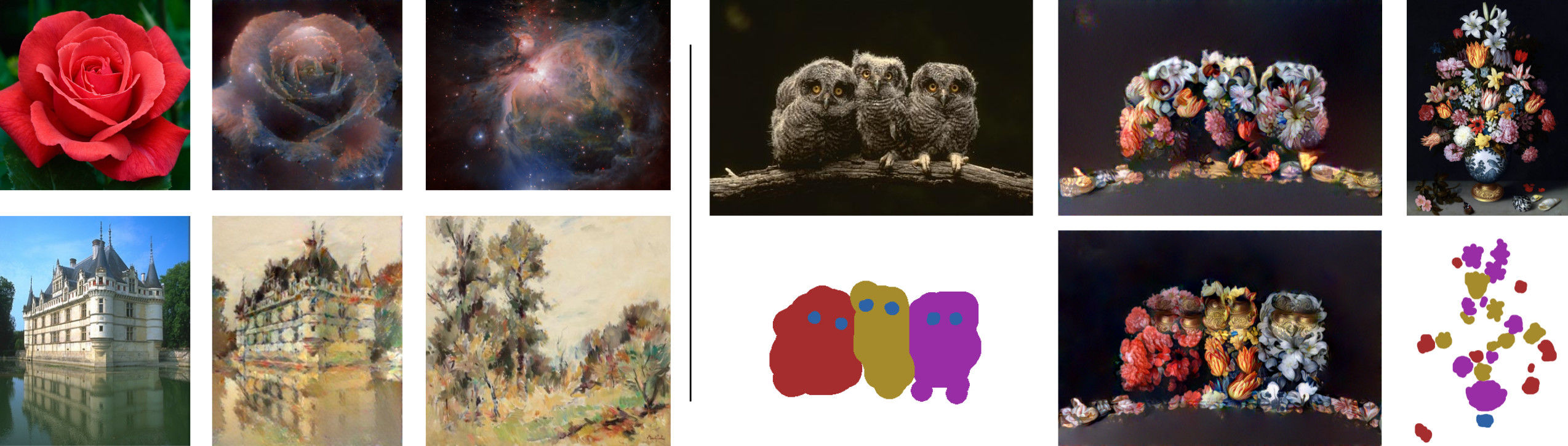}
	\caption{Examples of our output for unconstrained (left) and guided (right) style transfer.Images are arranged in order of content, output,
          style. Below the content and style image on the right we visualize the user-defined region-to-region guidance used to generate the output in the middle.}
	\label{fig:fig1}
\end{figure*}

\begin{abstract}
Style transfer algorithms strive to render the content of one image using the style of another. We propose Style Transfer by Relaxed Optimal Transport and Self-Similarity (STROTSS), a new optimization-based style transfer algorithm. We extend our method to allow user-specified point-to-point or region-to-region control over visual similarity between the style image and the output. Such guidance can be used to either achieve a particular visual effect or correct errors made by unconstrained style transfer. In order to quantitatively compare our method to prior work, we conduct a large-scale user study designed to assess the style-content tradeoff across settings in style transfer algorithms. Our results indicate that for any desired level of content preservation, our method provides higher quality stylization than prior work. Code is available \href{https://github.com/nkolkin13/STROTSS}{here}.
\end{abstract}

\input{Intro}

\input{Methods}

\input{Related}

\input{Results}

\input{Conclusion}

\clearpage
\bibliographystyle{abbrv}

\input{weakseg_bib}
\clearpage

\input{Appendix}

\clearpage 

\end{document}

%% file: Intro.tex
\cvprsection{Introduction}\label{sec:intro}
One of the main challenges of style transfer is formalizing
'content' and 'style', terms which evoke strong intuitions but 
are hard to even define semantically. We propose formulations of
each term which are novel in the domain of style transfer, but have a
long history of successful application in computer vision more
broadly. We hope that related efforts to refine definitions of 
both style and content will eventually lead to more robust recognition systems, but in this work we solely focus on their utility for style transfer.

We define style as a distribution over features extracted by
a deep neural network, and measure the distance between these
distributions using an efficient approximation of the Earth Movers
Distance initially proposed in the Natural Language Processing
community~\cite{kusner2015word}. This definition of style similarity
is not only well motivated statistically, but also intuitive. The goal
of style transfer is to deploy the visual attributes of the style image
onto the content image with minimum distortion to the content's underlying
layout and semantics; in essence to 'optimally transport' these visual attributes.

Our definition of content is inspired by the concept of
self-similarity, and the notion that human perceptual system is
robust because it identifies objects based on their appearance
relative to their surroundings, rather than absolute appearance. 
Defining content similarity in this way disconnects the term somewhat from
pixels precise values making it easier to satisfy than the definitions used in prior work. This allows the output of our algorithm to maintain the perceived semantics and spatial layout of the content image, while drastically differing in pixel space.

To increase utility of style transfer as an artistic tool, it is
important that users can easily and intuitively control the
algorithm's output. We extend our formulation to allow
region-to-region constraints on style transfer (e.g., ensuring that
hair in the content image is stylized using clouds in the style image)
and point-to-point constraints (e.g., ensuring that the eye in the
content image is stylized in the same way as the eye in a
painting). 

We quantitatively compare our method to prior work using human evaluations gathered from
662 workers on Amazon Mechanical Turk (AMT). Workers evaluated content preservation and stylization quality separately. Workers were shown two algorithms' output for the same inputs in addition to either the content or style input, then asked which has more similar 
content or style respectively to the displayed input. In this way are able to quantify the performance of each algorithm along both axes. By evaluate our method and prior work for 
multiple hyper-parameter settings, we also
measure the trade-off within each method between stylization and content preservation as 
hyper-parameters change.  Our results indicate that for any desired level of content 
preservation, our method provides higher quality stylization than prior work.

%% file: Methods.tex
\newcommand{\xfig}[1]{\includegraphics[height=1.6cm]{qual_cross/#1}}
\newcommand{\xfigLine}[1]{\xfig{#1} & \xfig{#1_01} & \xfig{#1_05} & \xfig{#1_03} & \xfig{#1_04} & \xfig{#1_02}}
\newcommand{\xfigLineS}[1]{ & \xfig{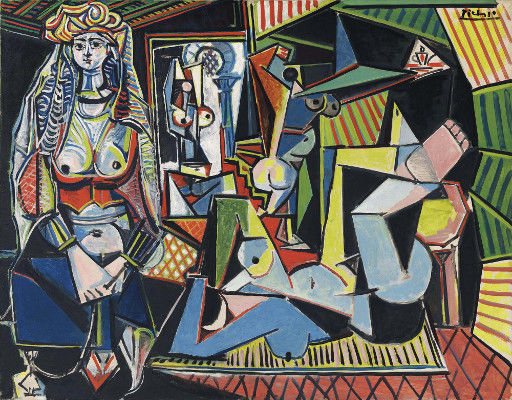} & \xfig{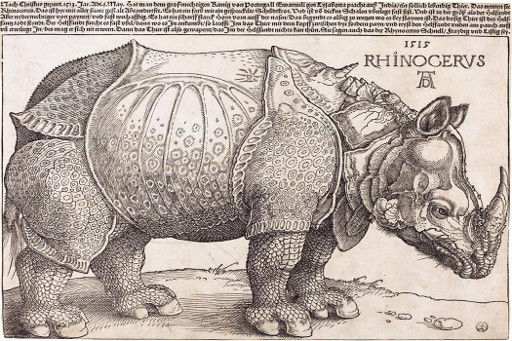} & \xfig{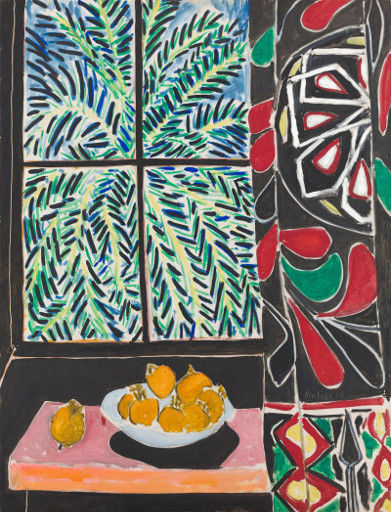} & \xfig{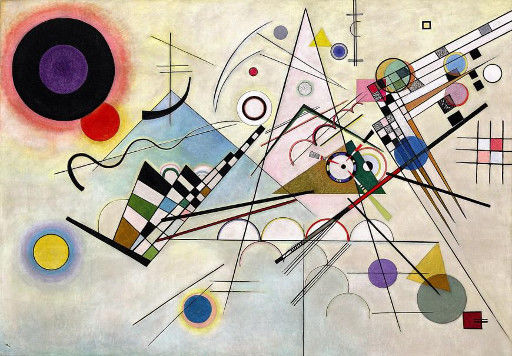} & \xfig{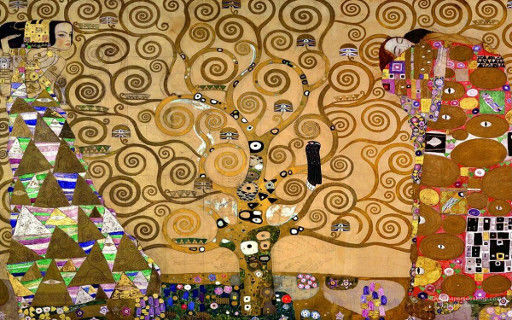}}

\begin{figure*}
	\centering
	\begin{tabular}{cccccc}
		\xfigLineS{01}\\
		\xfigLine{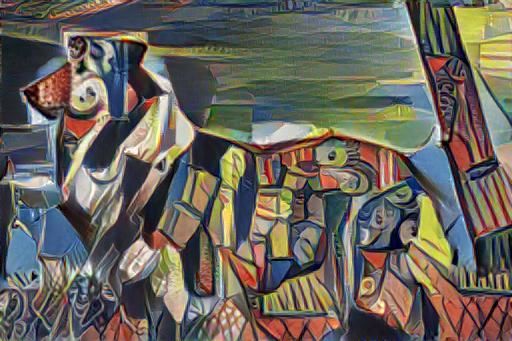}\\
		\xfigLine{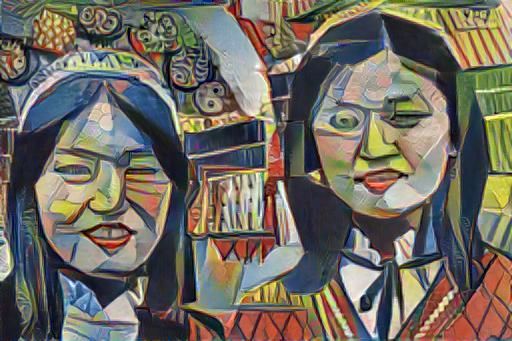}\\
		\xfigLine{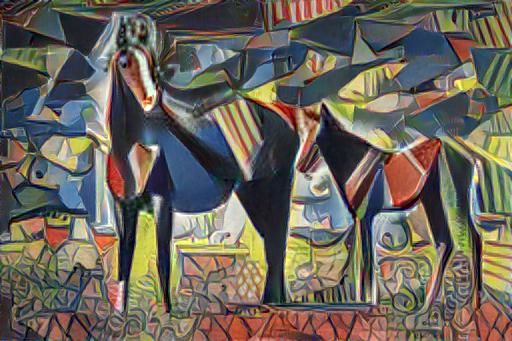}\\
		\xfigLine{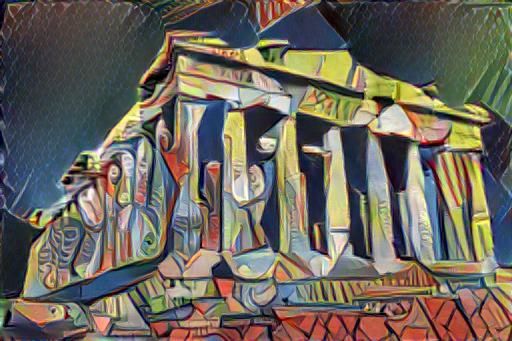}\\
		\xfigLine{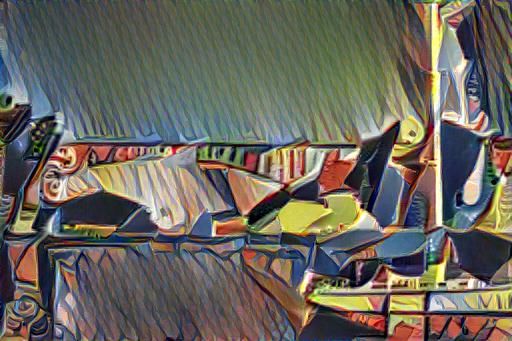}\\
		 & Picasso & D{\"u}rer & Matisse & Kandinsky & Klimt\\
	\end{tabular}
	\caption{Examples of the effect of different content images on the same style, and vice-versa}
	\label{fig:qual_cross}
\end{figure*}

\cvprsection{Methods}\label{sec:meth}
Like the original Neural Style Transfer algorithm proposed by Gatys et al.~\cite{gatys2016image} our method takes two inputs, a style image $I_S$ and a content image $I_C$, and uses the gradient descent variant RMSprop \cite{hinton2012neural} to minimize our proposed objective function (equation \ref{eq:full_obj}) with respect to the output image $X$.
\begin{align}
L(X,I_C,I_S)= \frac{\alpha\ell_C + \ell_m + \ell_r + \frac{1}{\alpha}\ell_p}{2+\alpha+\frac{1}{\alpha}}\label{eq:full_obj}
\end{align}

We describe the content term of our loss $\alpha \ell_C$ in Section \ref{sec:sl}, and the style term $\ell_m + \ell_r + \frac{1}{\alpha}\ell_p$ in  Section \ref{sec:cl}. The hyper-parameter $\alpha$ represents the relative importance of content preservation to stylization. Our method is iterative; let $X^{(t)}$ refer to the stylized output image at timestep $t$. We describe our initialization $X^{(0)}$ in Section \ref{subsec:impl}.

\cvprsubsection{Feature Extraction}
Both our style and content loss terms rely upon extracting a rich
feature representation from an arbitrary spatial location. In this
work we use hypercolumns \cite{mostajabi2015feedforward,
  hariharan2015hypercolumns} extracted from a subset of layers of VGG16 trained
on ImageNet~\cite{simonyan2014very}. Let
$\Phi(X)_i$ be the tensor of feature activations extracted from input
image $X$ by layer $i$ of network $\Phi$. Given the set of layer
indices $l_1,..,l_L$ we use bilinear upsampling to match the spatial
dimensions of $\Phi(X)_{l_1}...\Phi(X)_{l_L}$ to those of the original
image ($X$), then concatenate all such tensors along the feature dimension. This yields a hypercolumn at each pixel, that includes features which capture low-level edge and color features, mid-level texture features, and high-level semantic features \cite{zeiler2014visualizing}. For all experiments we use all convolutional layers of VGG16 except layers 9,10,12, and 13, which we exclude because of memory constraints.

\cvprsubsection{Style Loss}\label{sec:sl}
Let $A=\{A_1,\ldots,A_n\}$ be a set of $n$ feature vectors extracted
from $X^{(t)}$, and $B=\{B_1,\ldots,B_m\}$ be a set of $m$ features
extracted from style image $I_S$.
The style loss is derived from the Earth Movers Distance (EMD)\footnote{Since we consider
  all features to have equal mass, this is a simplified version
of the more general EMD~\cite{rubner1998metric}, which allows for
 transport between general, non-uniform mass distributions.}:
\begin{align}
\textsc{EMD}(A,B)=&\min_{T\geq 0} \sum_{ij} T_{ij}C_{ij}\\
s.t. &\sum_j T_{ij} = 1/m\label{eq:emda}\\
&\sum_i T_{ij} = 1/n\label{eq:emdb}
\end{align}
where $T$ is the 'transport matrix', which defines partial pairwise
assignments, and $C$ is the 'cost matrix' which defines how far an
element in $A$ is from an element in $B$. $\textsc{EMD}(A,B)$ captures
the distance between sets $A$ and $B$, but finding the optimal $T$ costs $O(\max(m,n)^3)$, and is therefore untenable for gradient descent based style
transfer (where it would need to be computed at each update
step). Instead we will use the Relaxed EMD~\cite{kusner2015word}. To
define this we will use two auxiliary distances, essentially each is
the EMD with only one of the constraints ~\eqref{eq:emda} or~\eqref{eq:emdb}:
\begin{align}
R_A(A,B)=&\min_{T\geq 0} \sum_{ij} T_{ij}C_{ij}\quad
s.t. &\sum_j T_{ij} = 1/m\\
R_B(A,B)=&\min_{T\geq 0} \sum_{ij} T_{ij}C_{ij}\quad
s.t. &\sum_i T_{ij} = 1/n
\end{align}
we can then define the relaxed earth movers distance as:
\begin{align}
\ell_r = REMD(A,B)=&\max (R_A(A,B),R_B(A,B))
\end{align}

This is equivalent to:
\begin{align}
\ell_r = &\max \left(\frac{1}{n}\sum_i \min_j C_{ij},\frac{1}{m}\sum_j \min_i C_{ij}\right)
\end{align}
Computing this is dominated by computing the cost matrix $C$. We compute the cost of transport (ground metric) from $A_i$ to $B_j$ as the
cosince distance between the two feature vectors,
\begin{align}
C_{ij}\,=\,D_{\cos}(A_i,B_j)\,=\,1-\frac{A_i \cdot B_j}{ \|A_i\| \|B_j\|}
\end{align}
We experimented with using the Euclidean distance between vectors instead, but the results were significantly worse, see the supplement for examples. 

While $\ell_r$ does a good job of transferring the structural forms of the source image to the target, the cosine distance ignores the magnitude of the feature vectors. In practice this leads to visual artifacts in the output, most notably over-/under-saturation. To combat this we add a moment matching loss:
\begin{equation}
\ell_m\,=\,\frac{1}{d}\|\mu_A-\mu_B\|_1 + \frac{1}{d^2}\|\Sigma_A-\Sigma_B\|_1
\end{equation}
where $\mu_A$, $\Sigma_A$ are the mean and covariance of the feature vectors in set $A$, and $\mu_B$ and $\Sigma_B$ are defined in the same way.

We also add a color matching loss, $\ell_p$ to encourage our output and the style image to have a similar palette.
$\ell_p$ is defined using the Relaxed EMD between pixel colors in
$X^{(t)}$ and $I_S$, this time and using Euclidean distance as a ground
metric. We find it beneficial to convert the colors from RGB into a decorrelated colorspace with mean color as one channel when computing this term. Because palette shifting is at odds with content preservation, we weight this term by $\frac{1}{\alpha}$. 

\begin{figure}[!b]
	\centering
	\includegraphics[width=0.4\textwidth]{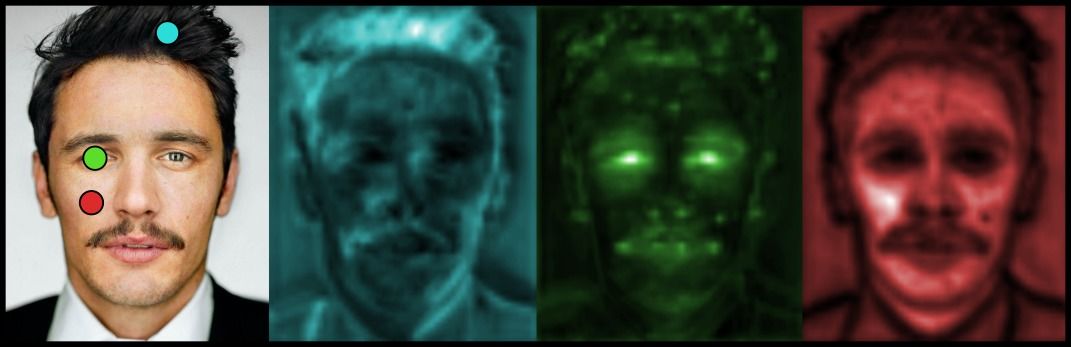}
	\caption{The blue, red, and green heatmaps visualize the cosine similarity in feature space relative to the corresponding points marked in the photograph. Our content loss attempts to maintain the relative pairwise similarities between 1024 randomly chosen locations in the content image}
	\label{fig:conSim}
\end{figure}

\cvprsubsection{Content Loss}\label{sec:cl}
Our content loss is motivated by the observation that robust pattern
recognition can be built using local self-similarity descriptors
\cite{shechtman2007matching}. An every day example of this is the
phenomenon called pareidolia, where the self-similarity patterns of
inanimate objects are perceived as faces because they match a loose
template. Formally, let $D^X$ be the pairwise cosine
distance matrix of all (hypercolumn) feature vectors extracted from $X^{(t)}$, and
let $D^{I_C}$ be defined analogously for the content image. We visualize several potential rows of $D^X$ in Figure \ref{fig:conSim}. We define our content loss as:
\begin{align}
\mathcal{L}_{content}(X,C) = \frac{1}{n^2} \sum_{i,j} \left|  \frac{D^X_{ij}}{\sum_i D^X_{ij}}- \frac{D^{I_C}_{ij}}{\sum_i D^{I_C}_{ij}}\right|
\end{align}
In other words the normalized cosine distance between feature vectors
extracted from any pair of coordinates should remain constant between
the content image and the output image. This constrains the structure
of the output, without enforcing any loss directly connected to pixels
of the content image. This causes the semantics and spatial layout to
be broadly preserved, while allowing pixel values in $X^{(t)}$ to
drastically differ from those in $I_C$.

\cvprsubsection{User Control}
We incorporate user control as constraints on the style of the output. Namely the user defines paired sets of spatial locations (regions) in $X^{(t)}$ and $I_S$ that must have low style loss. In the case of point-to-point user guidance each set contains only a single spatial location (defined by a click). Let us denote paired sets of spatial locations in the output and style image as $(X_{t1},S_{s1})...(X_{tK},S_{sK})$. We redefine the ground metric of the Relaxed EMD as follows:
\begin{align}
C_{ij} =
	\begin{cases}
	\beta*D_{cos}(A_i,B_j), \text{ if } i \in X_{tk}, j \in S_{sk} \\
	\infty, \text{ if } \exists k \text{ s.t. } i \in X_{tk} ,j \not \in S_{sk} \\
	D_{cos}(A_i,B_j) \text{ otherwise},
	\end{cases}
\end{align}
where $\beta$ controls the weight of user-specified constraints relative to the unconstrained portion of the style loss, we use $\beta=5$ in all experiments. In the case of point-to-point constraints we find it useful to augment the constraints specified by the user with 8 additional point-to-point constraints, these are automatically generated and centered around the original to form a uniform 9x9 grid. The horizontal and vertical distance between each point in the grid is set to be 20 pixels for 512x512 outputs, but this is is a tunable parameter that could be incorporated into a user interface.

\cvprsubsection{Implementation Details}\label{subsec:impl}
We apply our method iteratively at increasing resolutions, halving $\alpha$ each time. We begin with the content and style image scaled to have a long side of 64 pixels. The output at each scale is bilinearly upsampled to twice the resolution and
used as initialization for the next scale. By default we stylize at four resolutions, and because we halve $\alpha$ at each resolution our default $\alpha=16.0$ is set such that $\alpha=1.0$ at the final resolution.

At the lowest resolution we initialize using the bottom level of a Laplacian pyramid constructed from the content image (high frequency gradients) added to the mean color of the style image. We then decompose the initialized output image into a five level Laplacian pyramid, and use RMSprop~\cite{hinton2012neural} to update entries in the pyramid to minimize our objective function. We find that optimizing the Laplacian pyramid, rather than pixels directly, dramatically speeds up convergence. At each scale we make 200 updates using RMSprop, and use a learning rate of 0.002 for all scales except the last, where we reduce it to 0.001. 

The pairwise distance computation required to calculate the style and content loss precludes extracting features from all coordinates of the input images, instead we sample 1024 coordinates randomly from the style image, and 1024 coordinates in a uniform grid with a random x,y offset from the content image. We only differentiate the loss w.r.t the features extracted from these locations, and resample these locations after each step of RMSprop.

%% file: Related.tex
\newcommand{\qCIm}[3]{\includegraphics[height=#3\textwidth]{Figures/compare/#1_#2} }

\newcommand{\qCRow}[2]{\qCIm{#1}{Content}{#2} & \qCIm{#1}{Style}{#2} & \qCIm{#1}{Ours}{#2} & \qCIm{#1}{Reshuffle}{#2} & \qCIm{#1}{Gatys}{#2} & \qCIm{#1}{CNNMRF}{#2} & \qCIm{#1}{Contextual}{#2} }

\begin{figure*}
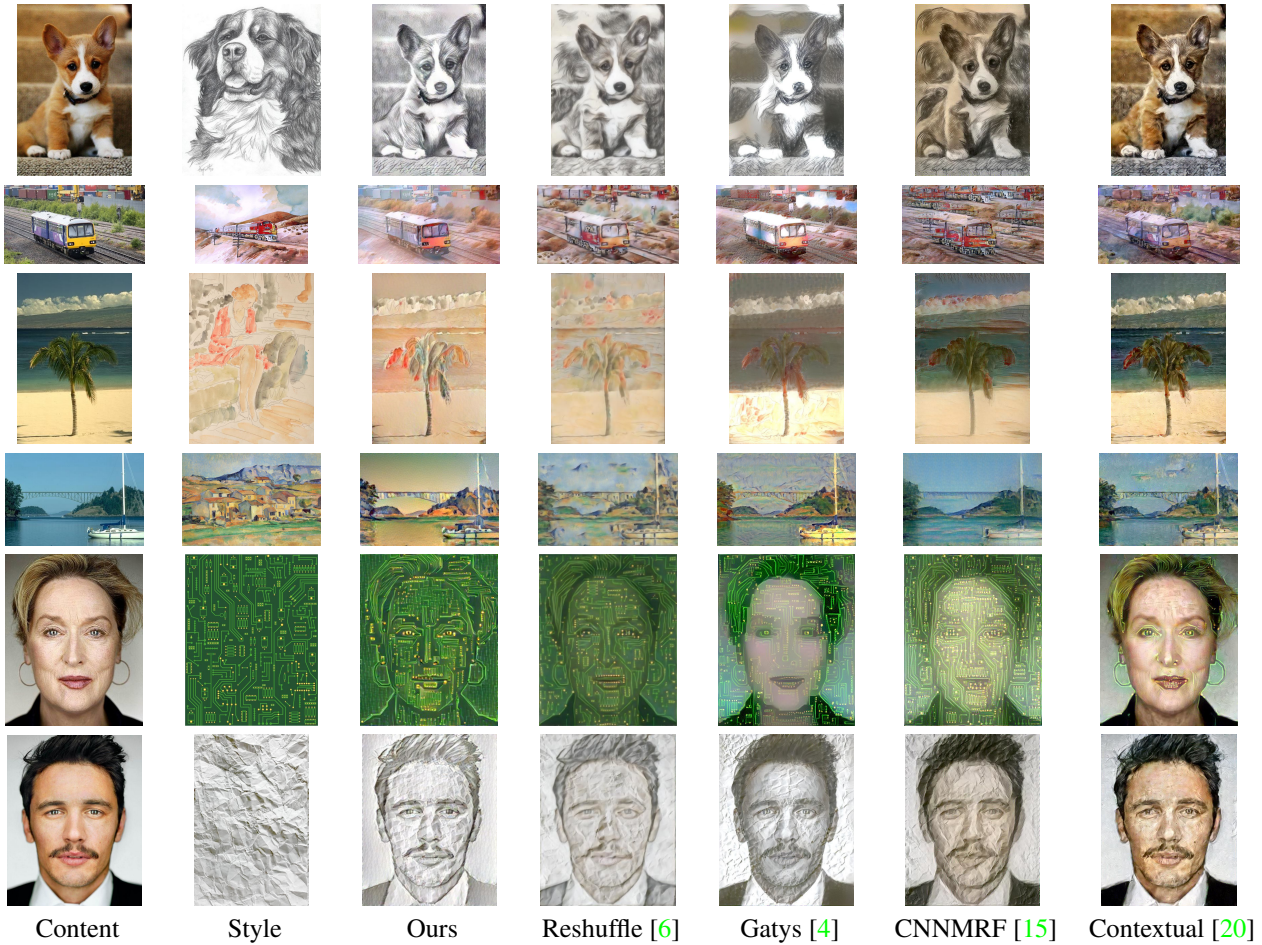

	\centering
	
	\begin{tabular}{ccccccc}
	\qCRow{00}{0.13}\\
	\qCRow{01}{0.06}\\
	\qCRow{02}{0.13}\\
	\qCRow{03}{0.07}\\
	\qCRow{04}{0.13}\\
	\qCRow{05}{0.13}\\
	Content & Style & Ours & Reshuffle \cite{gu2018arbitrary} & Gatys \cite{gatys2016image} & CNNMRF \cite{li2016combining} & Contextual \cite{mechrez2018contextual} 
\end{tabular}

	\caption{Qualitative comparison between our method and prior work. Default hyper-parameters used for all methods}
	\label{fig:comp_fig}
\end{figure*}

\cvprsection{Related Work}\label{sec:related}

Style transfer algorithms have existed for decades, and traditionally
relied on hand-crafted algorithms to render an image in fixed style
\cite{haeberli1990paint,hertzmann1998painterly}, or 
hand-crafting features to be matched between an arbitrary style to the content image
\cite{hertzmann2001image,efros2001image}. The
state-of-the-art was dramatically altered in 2016 when Gatys et
al.~\cite{gatys2016image} introduced Neural Style Transfer. This
method uses features extracted from a neural network pre-trained for
image classification. It defines style in terms of the Gram matrix of
features extracted from multiple layers, and content as the feature
tensors extracted from another set of layers. The style loss is
defined as the Frobenius norm of the difference in Gram
feature matrices between the output image and style image. The content
loss is defined as the Frobenius norm of the difference between
feature tensors from the output image and the style image. Distinct
from the framework of Neural Style Transfer, there are several recent
methods \cite{liao2017visual, Aberman2018} that use similarities
between deep neural features to build a correspondence map between the
content image and style image, and warp the style image onto the
content image. These methods are extremely successful in paired
settings, when the
contents of the style image and content image are similar, but are not
designed for style transfer between arbitrary images (unpaired or
texture transfer).

Subsequent work building upon \cite{gatys2016image} has explored
improvements and modifications along many axes. Perhaps the most
common form of innovation is in proposals for quantifying the 'stylistic similarity' between two images~\cite{li2016combining, berger2016incorporating, risser2017stable, mechrez2018contextual}. For example in order to capture long-range spatial dependencies Berger et al.~\cite{berger2016incorporating} propose computing multiple Gram matrices using translated feature tensors (so that the outer product is taken between feature vectors at fixed spatial offsets). Both~\cite{gatys2016image} and~\cite{berger2016incorporating} discard valuable information about the complete distribution of style features that isn't captured by Gram matrices.

In~\cite{li2016combining} Li et al. formulate the style loss as
minimizing the energy function of a Markov Random Field over the
features extracted from one of the latter layers of a pretrained CNN,
encouraging patches (which yielded the deep features) in the target
image to match their nearest neighbor from style image in feature
space. Other functionally similar losses appear
in~\cite{risser2017stable}, which treats style transfer as matching
two histograms of features, and~\cite{mechrez2018contextual}, which
matches features between the style and target which are significantly
closer than any other pairing. In all of these methods, broadly speaking,
features of the output are encouraged to
lie on the support of the distribution of features extracted from the
style image, but need not cover it. These losses are all similar to
one component of the Relaxed EMD ($R_A$). However, our method differs
from these approaches because our style term also encourages covering
the entire distribution of features in the style image ($R_B$). Our
style loss is most similar in spirit to that proposed by Gu et
al~\cite{gu2018arbitrary}, which also includes terms that encourage
fidelity and diversity. Their loss minimizes the distance
between explicitly paired individual patches, whereas ours minimizes the distance between
distributions of features. 

Another major category of innovation is replacing the
optimization-based algorithm of~\cite{gatys2016image} with a
neural network trained to perform style transfer, enabling real-time
inference. Initial efforts in this area were constrained to a limited
set of pre-selected styles~\cite{johnson2016perceptual}, but
subsequent work relaxed this constraint and allowed arbitrary styles
at test time~\cite{huang2017arbitrary}. Relative to slower
optimization-based methods these works made some sacrifices in the
quality of the output for speed. However, Sanakoyeu et
al.~\cite{sanakoyeu2018styleaware} introduce a method for
incorporating style images from the same artist into the real-time
framework which produces high quality outputs in real-time, but in
contrast to our work relies on having access to multiple images with
the same style and requires training the style transfer mechanism separately for each new style.

Various methods have been proposed for controlling the output of style
transfer. In~\cite{gatys2017controlling} Gatys et al. propose two
'global' control methods, that affect the entire output rather than a
particular spatial region. One method is decomposing the image into
hue, saturation, and luminance, and only stylizes the luminance in
order to preserve the color palette of the content image. A second
method from~\cite{gatys2017controlling} is to generate an auxiliary
style image either to preserve color, or to transfer style from only a
particular  scale (for example the transferring only the brush-strokes, rather than the larger and more structurally
complex elements of the style). These types of user control are
orthogonal to our method, and can be incorporated into it.

Another type of control is spatial, allowing users to ensure that certain regions of the output should be stylized using only features from a manually selected region of the style image (or that different regions of the output image should be stylized based on different style images). In~\cite{gatys2017controlling, lu2017decoder} the authors propose forms of spatial control based on the user defining matched regions of the image by creating a dense mask for both the style and content image. We demonstrate that it is straightforward to incorporate this type of user-control into our formulation of style transfer. In the supplement we show an example comparing the spatial control of our method and ~\cite{gatys2017controlling}, and demonstrate that both yield visually pleasing results that match the spatial guidance provided.

\begin{figure*}[!tbh]
	\centering
	\includegraphics[width=0.9\textwidth]{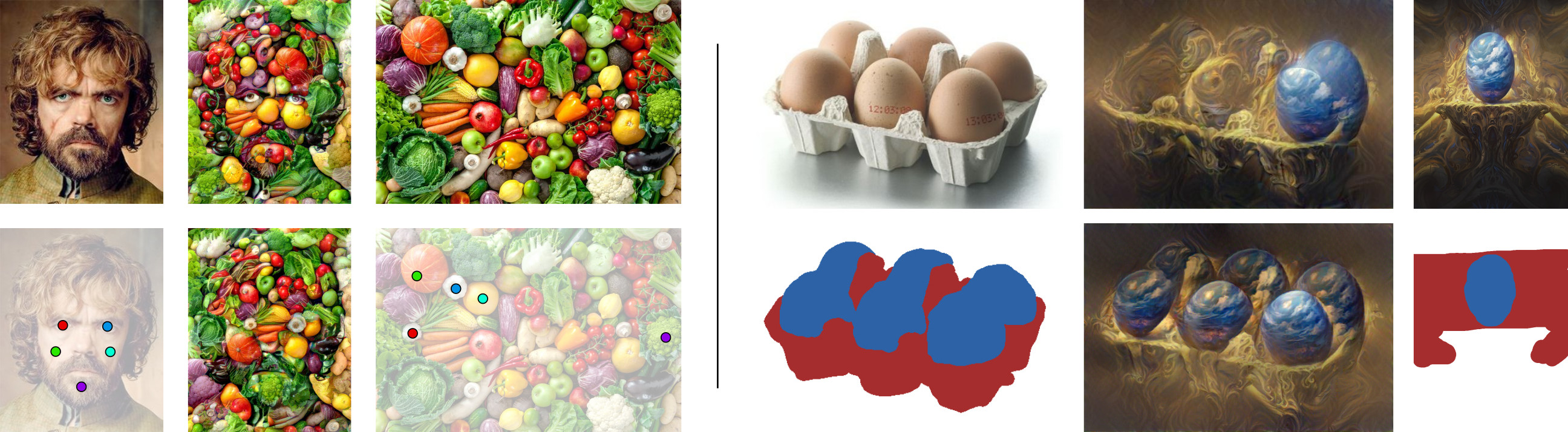} 
	\caption{Examples of using guidance for
          aesthetic effect (left, point-to-point)) and error
          correction (right, region-to-region). In the top row the
          images are arranged in order of content, output,
          style. Below each content and style image we show the
          guidance mask, and between them the guided output.}
	\label{fig:control_fig}
\end{figure*}

\newcommand{\cKIm}[3]{\includegraphics[height=#3cm]{content_knob/#1_#2}}

\newcommand{\contentKnobRow}[2]{\cKIm{#1}{c}{#2} & \cKIm{#1}{2000}{#2} & \cKIm{#1}{1000}{#2} & \cKIm{#1}{0500}{#2} & \cKIm{#1}{0250}{#2} & \cKIm{#1}{s}{#2}}

\begin{figure*}[tb]
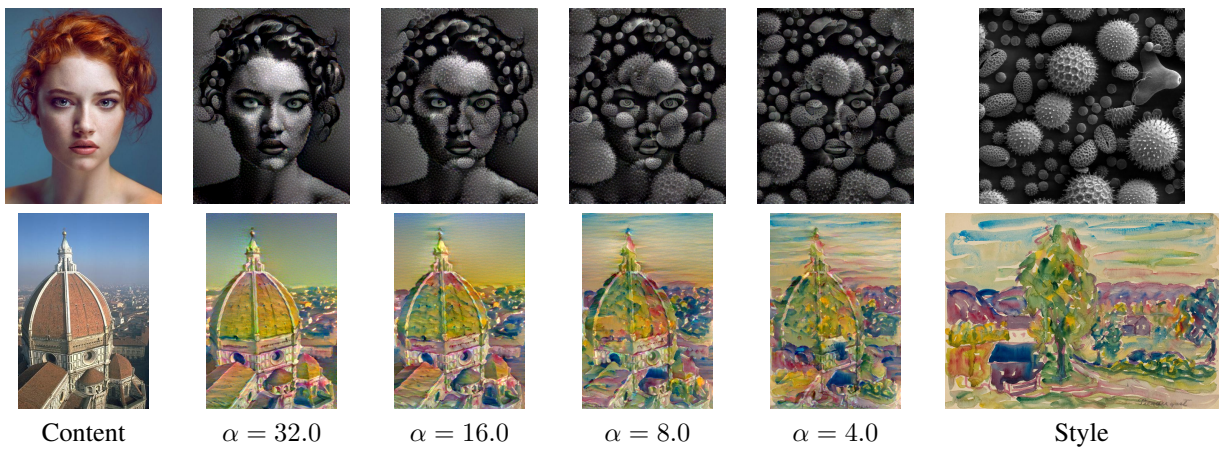

	\centering
\begin{tabular}{cccccc}
	\contentKnobRow{pollen}{2.6}\\
	\contentKnobRow{dome}{2.6}\\
	Content & $\alpha=32.0$ & $\alpha=16.0$ & $\alpha=8.0$ & $\alpha=4.0$ & Style
\end{tabular}\vspace{-.5em}
	\caption{Effect of varying $\alpha$, the content loss weight, on our unconstrained style transfer output, because we stylize at four resolutions, and halve $\alpha$ each time, our default $\alpha=16.0$ is set such that $\alpha=1.0$ at the final resolution.}
	\label{fig:knob_fig}
\end{figure*}

Evaluating and comparing style transfer algorithms is a challenging
task because, in contrast to object recognition or segmentation, there
is no established ``ground truth'' for the output. The most common
method is a qualitative, purely subjective comparison between the
output of different algorithms. Some methods also provide more refined
qualitative comparisons such as texture
synthesis~\cite{risser2017stable,gu2018arbitrary} and
inpainting~\cite{berger2016incorporating}. While these comparisons
provide insight into the behavior of each algorithm, without
quantitative comparisons it is difficult to draw conclusions about the
algorithm's performance on average. The most common quantitative
evaluation is asking users to rank the output of each algorithm
according to aesthetic appeal~\cite{gu2018arbitrary, li2018closed,
  mechrez2017photorealistic}. Recently Sanakoyeu et
al.~\cite{sanakoyeu2018styleaware} propose two new forms of
quantitative evaluation. The first is testing if an neural network
pretrained for artist classification on real paintings can correctly
classify the artist of the style image based on an algorithm's
output. The second is asking experts in art history which algorithm's
output most closely matches the style image. We designed our human
evaluation study, described in section \ref{sec:h_eval}, to give a
more complete sense of the trade-off each algorithm makes between
content and style as its hyper-parameters vary. To the best of our knowledge it is the first such effort.

%% file: Results.tex
\begin{figure}[!bth]
	\centering
	\includegraphics[width=0.45\textwidth]{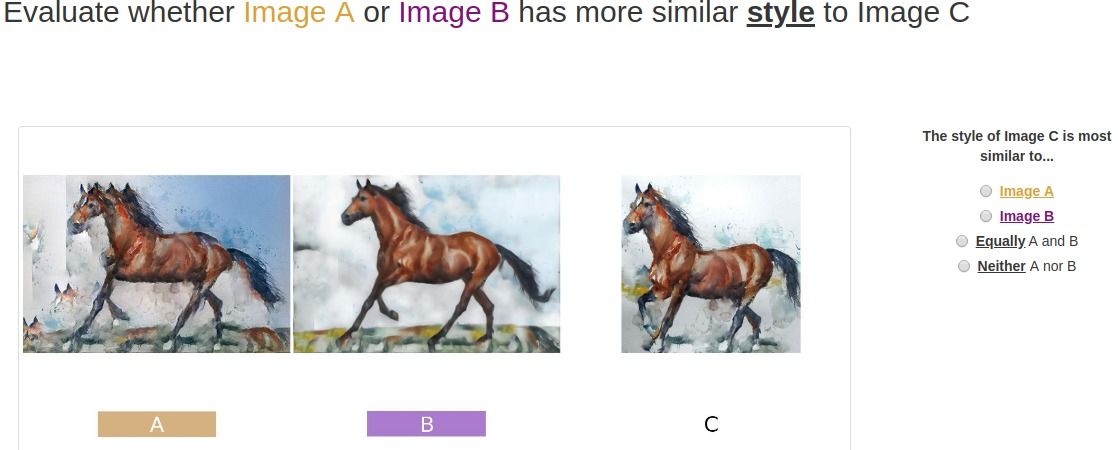}
	\caption{Human evaluation interface}
	\label{fig:mturk_gui}\vspace{-1.5em}
\end{figure}

\cvprsection{Experiments}
We include representative qualitative results in Figures~\ref{fig:qual_cross},~\ref{fig:comp_fig},
and an illustration of the effect of the content weight $\alpha$ in
Figure~\ref{fig:knob_fig}. Figure~\ref{fig:control_fig} demonstrates
uses of user guidance with our method.

\cvprsubsection{Large-Scale Human Evaluation}\label{sec:h_eval}
Because style transfer between arbitrary content and style pairs is such a broad task, we propose three regimes that we believe cover the major use cases of style transfer. 'Paired' refers to when the content image and style image are both representations of the same things, this is mostly images of the same category (e.g. both images of dogs), but also includes images of the same entity (e.g. both images of the London skyline). 'Unpaired' refers to when the content and style image are \textbf{not} representations of the same thing (e.g. a photograph of a Central American temple, and a painting of a circus). 'Texture' refers to when the content is a photograph of a face, and the style is a homogeneous texture (e.g. a brick wall, flames). For each regime we consider 30 style/content pairings (total of 90).

\begin{figure*}[!tbh]
	\centering
	\includegraphics[width=1.0\textwidth]{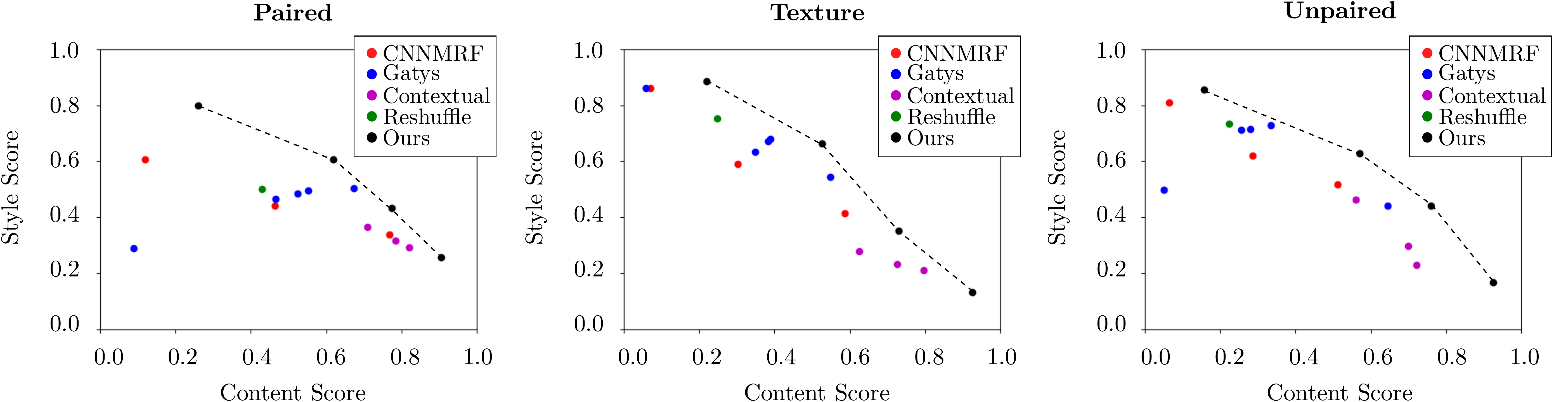}
	\caption{Quantitative evaluation of our method and prior work, we estimate the Pareto frontier of the methods evaluated by linearly interpolation (dashed line)}
	\label{fig:mturk_results_fig}
\end{figure*}

In order to quantitatively compare our method to prior work we
performed several studies using AMT. An example of the workers' interface is shown in
Figure~\ref{fig:mturk_gui}. Images A and B were the result of the same
inputs passed into either the algorithms proposed
in~\cite{gatys2016image},\cite{gu2018arbitrary},~\cite{li2016combining},~\cite{mechrez2018contextual},
or our method. In Figure~\ref{fig:mturk_gui} image C is the corresponding
style image, and workers were asked to choose whether the style of
image is best matched by: 'A', 'B', 'Both Equally', or 'Neither'. If
image C is a content image, workers are posed the same question with
respect to content match, instead of style. For each competing
algorithm except \cite{gu2018arbitrary} we test three sets of
hyper-parameters, the defaults recommended by the authors, the same
with $\frac{1}{4}$ of the content weight (high stylization), and the
same with double the content weight (low stylization). Because these modifications to content weight did not alter the behavior of ~\cite{gatys2016image} significantly we also tested ~\cite{gatys2016image} with $\frac{1}{100}$ and $100\times$ the default content weight. We also test our method with $4\times$ the content weight. We only were able to test the default hyper-parameters for~\cite{gu2018arbitrary} because
the code provided by the authors does not expose content weight as a
parameter to users. We test all possible pairings of A and B between
different algorithms and their hyperparameters (i.e. we do not compare
an algorithm against itself with different hyperparameters, but do
compare it to all hyperparameter settings of other algorithms). In
each presentation, the order of output (assignment of methods to A or B in
the interface) was randomized. Each pairing was voted on by an average
of 4.98 different workers (minimum 4, maximum 5), 662 workers in
total. On average, 3.7 workers agreed with the majority vote for each
pairing. All of the images used in this evaluation will be made
available to enable further benchmarking.

For an algorithm/hyper-parameter combination we define its content score to be the number of times it was selected by workers as having closer or equal content to $I_C$ relative to the other output it was shown with, divided by the total number of experiments it appeared in. This is always a fraction between 0 and 1. The style score is defined analogously. We present these results in Figure \ref{fig:mturk_results_fig}, separated by regime. The score of each point is computed over 1580 pairings on average (including the same pairings being shown to distinct workers, minimum 1410, maximum 1890). Overall for a given level of content score, our method provides a higher style score than prior work.

\cvprsubsection{Ablation Study}

\newcommand{\ablationrow}[2]{\includegraphics[height=#2\textwidth]{Figures/textureSyn/#1_moment} & \includegraphics[height=#2\textwidth]{Figures/textureSyn/#1_remd1} & \includegraphics[height=#2\textwidth]{Figures/textureSyn/#1_remd2} & \includegraphics[height=#2\textwidth]{Figures/textureSyn/#1_remd} & \includegraphics[height=#2\textwidth]{Figures/textureSyn/#1_MomRemd} & \includegraphics[height=#2\textwidth]{Figures/textureSyn/#1_standard} & \includegraphics[height=#2\textwidth]{Figures/textureSyn/#1}}

\begin{figure*}[!bth]
	\centering
	\begin{tabular}{ccccccc}
	\ablationrow{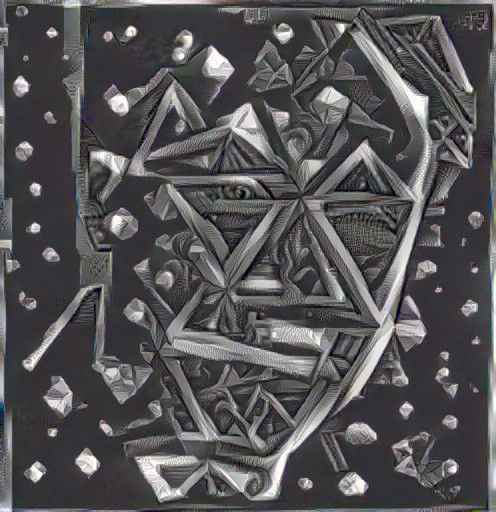}{0.13} \\
	$\ell_{m}$ & $\ell_{R_A}$ & $\ell_{R_B}$ & $\ell_{r}$ & $\ell_{r}$ + $\ell_{m}$ & $\ell_{r}$ + $\ell_{m}$ + $\frac{\ell_{p}}{\alpha}$ & Style \\
	\end{tabular}
	\caption{Ablation study of effects of our proposed style terms with low content loss ($\alpha=4.0$). See text for analysis of each terms' effect. Best viewed zoomed-in on screen.}
	\label{fig:mturk_ab_fig}
\end{figure*}

In Figure \ref{fig:mturk_ab_fig} we explore the effect of different
terms of our style loss, which is composed of a moment-matching loss
$\ell_m$, the Relaxed Earth Movers Distance $\ell_r$, and a color
palette matching loss $\ell_p$. As seen in Figure
\ref{fig:mturk_ab_fig}, $\ell_m$ alone does a decent job of
transferring style, but fails to capture the larger structures of the style image. $\ell_{R_A}$ alone does not make use of the entire distribution of style features, and reconstructs content more poorly than $\ell_{r}$. $\ell_{R_B}$ alone encourages every style feature to have a nearby output feature, which is too easy to satisfy. Combining $\ell_{R_A}$ and $\ell_{R_B}$ in the relaxed earth movers distance $\ell_r$ results in a higher quality output than either term alone, however because the ground metric used is the cosine distance the magnitude of the features is not constrained, resulting in saturation issues. Combining $\ell_{r}$ with $\ell_{m}$ alleviates this, but some issues with the output's palette remain, which are fixed by adding $\ell_p$.


\cvprsubsection{Relaxed EMD Approximation Quality}\label{subsec:approx}
To measure how well the Relaxed EMD approximates the exact Earth
Movers Distance we take each of the 900 possible
content/style pairings formed by the 30 content and style images used in our
AMT experiments for the unpaired regime. For each pairing we compute the REMD between 1024 features
extracted from random coordinates, and the exact EMD
based on the same set of features.
We then analyze the distribution of $\frac{REMD(A,B)}{EMD(A,B}$
Because the REMD is a lower bound, this quantity is always
$\le$1. Over the 900 image pairs, its mean was 0.60, with standard deviation
0.04. A better EMD approximation, or
one that is an upper bound rather than a lower bound, may yield better
style transfer
results. On the other hand the REMD is simple to compute, empirically easy to optimize, and yields good results.

\cvprsubsection{Timing Results}
We compute our timing results using a Intel i5-7600 CPU @ 3.50GHz CPU,
and a NVIDIA GTX 1080 GPU. We use square style and content images
scaled to have the edge length indicated in the top row of Table
\ref{tab:timing}. For inputs of size 1024x1024 the methods from
\cite{li2016combining} and \cite{mechrez2018contextual} ran out of
memory ('X' in the table). Because the code provided by the authors \cite{gu2018arbitrary} only runs on Windows, we had to run it on a different computer. To approximate the speed of their method on our hardware we project the timing result for 512x512 images reported in their paper based on the relative speedup for \cite{li2016combining} between their hardware and ours. For low resolution outputs our method is relatively slow, however it scales better for outputs with resolution 512 and above relative to \cite{li2016combining} and \cite{mechrez2018contextual}, but remains slower than \cite{gatys2016image} and our projected results for \cite{gu2018arbitrary}.

\begin{table}[t]
\centering
\begin{tabular}{c||c|c|c|c|c}
	\textbf{Image size} & \textbf{64} & \textbf{128} & \textbf{256} & \textbf{512} & \textbf{1024}\\
	\hline \hline
	Ours & 20 & 38 & 60 & 95 & 154\\
	Gatys & 8 & 10 & 14 & 33 & 116\\
	CNNMRF & 3 & 8 & 27 & 117 & X\\
	Contextual & 13 & 40 & 189 & 277 & X\\
	Reshuffle & - & - & - & \textit{69*} & -\\
\end{tabular}
	\caption{Timing comparison (in seconds) between our methods
          and others. The style and content images had the same
          dimensions and were square. *: a projected
          result, see text for details. -: we were not able to project these results. X: the method ran out of memory.}
	\label{tab:timing}
\end{table}

%% file: Conclusion.tex
\cvprsection{Conclusion and Future Work}
We propose novel formalizations of style and content for style transfer and show that the resulting algorithm compares favorably to prior work, both in terms of stylization quality and content preservation. Via our ablation study we show that style-similarity losses which more accurately measure the distance between distributions of features leads to better style transfer. The approximation of the earth movers distance that we use is simple, but effective, and we leave it to future work to explore more accurate approximations. Another direction for future work is improving our method's speed by training feed-forward style transfer methods using our proposed objective function.

%% file: Appendix.tex
\cvprsection{Appendix}

\begin{figure*}[ht]
\includegraphics[width=\textwidth]{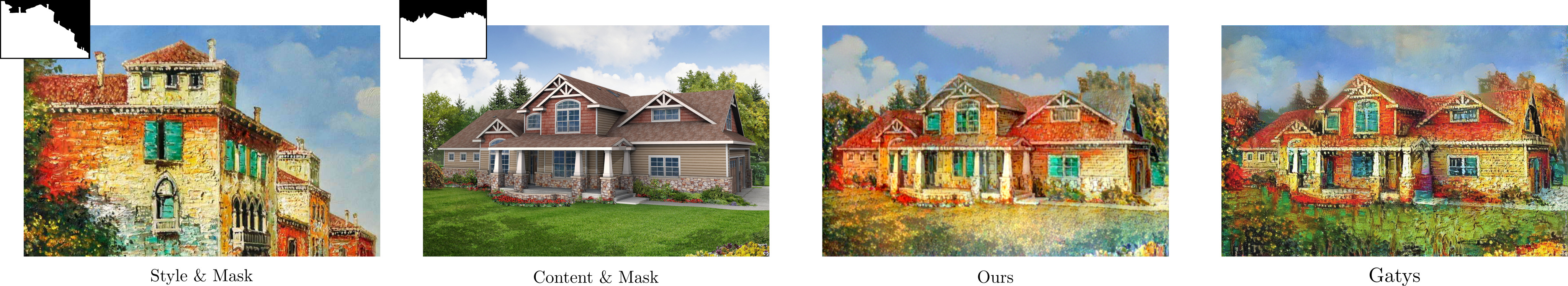}
	\caption{Qualitative comparison of the resulting output of our spatial guidance and that proposed in ~\cite{gatys2017controlling}}
	\label{fig:mturk_guid_comp_fig}
\end{figure*}

In order to demonstrate that our proposed method for spatial guidance gives the same level of user-control as those previously proposed we provide a qualitative comparison in Figure \ref{fig:mturk_guid_comp_fig}. For the same content and style with the same guidance masks we show the output of our method, and the output of the method proposed in ~\cite{gatys2017controlling} using one of the examples from their paper.

In Figure \ref{fig:mturk_ab_fig} we show an extended ablation study of our method. For each content image and style image we show the effect of different style losses or algorithmic decisions on our output. 'Optimize Pixels' refers to performing gradient descent on pixel values of the output directly, instead of the entries of a laplaccian pyramid (our default). In 'Single Scale' we perform 800 updates at the final resolution, instead of 200 updates at each of four increasing resolutions. In '$\ell_2$ Ground Metric' we replace the ground metric of the Relaxed EMD with euclidean distance (instead of our default, cosine distance). The other style loss ablations are explained in Section 4.2 of the main text.

\newcommand{\ablationrowApp}[2]{
\includegraphics[height=#2\textwidth]{Figures/textureSyn/#1_c} &\includegraphics[height=#2\textwidth]{Figures/textureSyn/#1}& \includegraphics[height=#2\textwidth]{Figures/textureSyn/#1_standard} & \includegraphics[height=#2\textwidth]{Figures/textureSyn/#1_pixDirect}&\includegraphics[height=#2\textwidth]{Figures/textureSyn/#1_singleScale}&\includegraphics[height=#2\textwidth]{Figures/textureSyn/#1_l2Ground} \\
Content & Style & $\ell_{r}$ + $\ell_{m}$ + $\frac{\ell_{p}}{\alpha}$ & Optimize Pixels & Single Scale & $\ell_2$ Ground Metric\\
&\includegraphics[height=#2\textwidth]{Figures/textureSyn/#1_moment} & \includegraphics[height=#2\textwidth]{Figures/textureSyn/#1_remd1} & \includegraphics[height=#2\textwidth]{Figures/textureSyn/#1_remd2} & \includegraphics[height=#2\textwidth]{Figures/textureSyn/#1_remd} & \includegraphics[height=#2\textwidth]{Figures/textureSyn/#1_MomRemd}\\
&$\ell_{m}$ & $\ell_{R_A}$ & $\ell_{R_B}$ & $\ell_{r}$ & $\ell_{r}$ + $\ell_{m}$}

\begin{figure*}[hb]
	\centering
	\begin{tabular}{ccccccc}
	\ablationrowApp{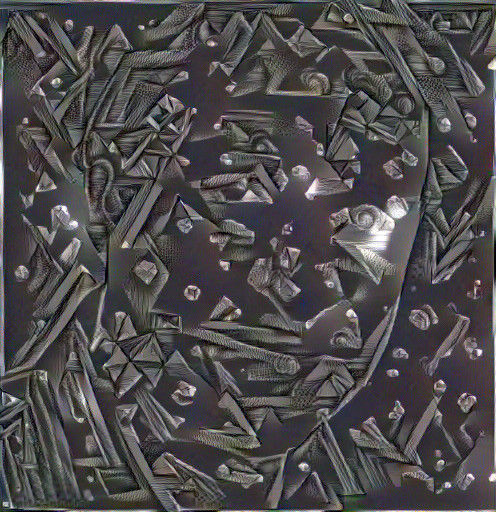}{0.13} \\
	\ablationrowApp{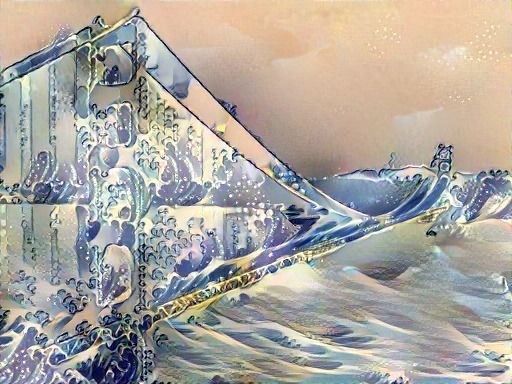}{0.09} \\
	\ablationrowApp{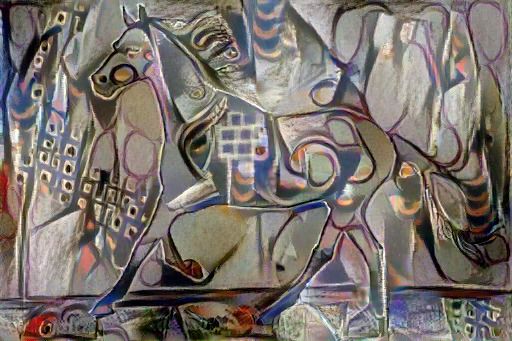}{0.09} \\
	\end{tabular}
	\caption{Extended ablation study}
	\label{fig:mturk_ab_fig}
\end{figure*}